\pdfoutput=1
\pdfoutput=1
\pdfoutput=1
\pdfoutput=1
\pdfoutput=1
\documentclass[journal=jacsat,manuscript=article]{achemso}

\usepackage[version=3]{mhchem} 

\author{Fardin Ghorbani}
\author{Javad Shabanpour}
\email{m.javadshabanpour1372@gmail.com}
\affiliation[Iran University of Science and Technology]
{Department of Electrical Engineering, Iran University of Science and Technology, Narmak, Tehran 16486-13114, Iran}
\author{Sina Beyraghi}
\author{Hossein Soleimani}
\author{Homayoon Oraizi}
\author{Mohammad Soleimani}
\title[An \textsf{achemso} demo]
  {A deep learning approach for inverse design of the metasurface for dual-polarized waves}

\abbreviations{IR,NMR,UV}

\begin{document}
\begin{abstract}
	Compared to the conventional metasurface design, machine learning-based
	methods have recently created an inspiring platform for an inverse realization of the metasurfaces. Here, we have used the Deep Neural Network (DNN) for the generation of desired output unit cell structures for both TE and TM polarized
	waves which its working frequency can reach up to 45 GHz. To automatically generate metasurfaces over wide frequencies, we deliberately design 8 annular models; thus, each generated meta-atoms in our dataset can produce different notches in our desired working frequency. Compared to the general approach, whereby the
	final metasurface structure may be formed by any randomly distributed ``0" and ``1", we
	propose here a confined output configuration. By confining the output, the number 
	of calculations will be decreased and the learning speed will be increased. 
	Establishing a DNN-confined output configuration based on the input data for both TE and TM polarized waves is the novelty to generate the desired metasurface structure for dual orthogonal polarizations.
	Moreover, we have demonstrated that our network can attain an accuracy of 92\%.
	Obtaining the
	final unit cell directly without any time-consuming optimization algorithms for both
	TE and TM polarized waves, and high average accuracy, open beneficial ways for the inverse metasurface design; thus, the designer is required only to focus on the design goal.
\end{abstract}

\maketitle

\section{Introduction}
Metamaterials have attracted widespread attention due to their peculiar assets to modify the permittivity and permeability \cite{1,2,27,28,29}. Recently, many novel functionalities have been implemented by metamaterials and their 2D counterpart, metasurfaces, such as intelligent surfaces for communication\cite{3,4}, real-time wavefront manipulation\cite{5,6,7,8}, perfect absorption\cite{9,10}, and machine learning metasurface design\cite{11,12}.

However, all of these works are founded on conventional approaches including trial-and-error methods, brute force optimization methods, and parameter sweep, which are time-consuming processes. Therefore, to solve the above challenges and to seek a fast, effective, and programmed route for designing a metasurface, we have benefited from machine learning. Deep learning is an efficient approach for learning the relationship between input and wanted information from the samples of past experiences. To be more specific, deep learning as a special section of machine learning can infer the basic rules based on formerly specified data; then, for different assigned inputs, the designed network can estimate reasonable decisions. With expanding development of machine learning and its possible future applications to address some important problems such as signal processing\cite{20} and through the wall imaging\cite{21}, we are now witnessing the opening of machine learning in wave-interaction phenomena. Owing to its potential capacity to provide higher accuracy, less design time, and enhance the productivity of a modeling procedure, machine learning has been introduced in numerous electromagnetic phenomena, for instance, all-dielectric metasurfaces\cite{13}, antenna design\cite{14,15}, acoustic metamaterials\cite{16,17}, and computational electromagnetics\cite{18,19}.  

T. Cui et al. presented an inverse metasurface realization that can recognize the internal principles between input metasurface structures and their S-parameters with 76.5\% of accuracy.\cite{22}. Zhang et al. have proposed an approach based on machine learning for designing anisotropic coding metasurfaces with a connection between deep learning and BPSO to explore the ideal reflection phases of two-unit cells for the desired target\cite{23}. Shan et al. have introduced conditional deep convolutional
generative adversarial networks to code the programmable metasurface for multiple beam steering, with accuracy higher than 94\%.\cite{24}. A machine learning-based inverse metasurface design has been provided, in\cite{25} which is capable of directly computing the output metasurface structure by entering the sought design targets into the model.
Recently, a double deep Q-learning network has been introduced for distinguishing the ideal type of material and optimizing a hologram structure to increase its efficiency.\cite{26}

Here, benefiting from Deep Neural Network (DNN), we propose a network scheme for automatic metasurface design for dual-polarized waves with an average accuracy of up to 92\%.
Our network is capable of generating any desired metasurface based on the input data for both TE and TM polarized waves which its working frequency can reach up to 45 GHz. The works presented in \cite {22,25} can generate the output structure in the frequency range of 5 to 20 GHz and 16-20 GHz, respectively. To broaden the working frequency, we consider 8 annular models with the purpose of generating single or multiple resonances in the desired working frequency. Besides, to enhance the speed of the training, shorten the number of computations, and boost the effectiveness of our network, the output of the network is confined; thus, the DNN should generate the metasurface structure by employing the proposed 8 annular models. We demonstrate that the accuracy of this network also reaches 91\%. Consuming less computational resources, having a wide frequency band, generating desired output metasurface without resorting to optimization procedure, and working for both TE and TM polarized waves make our method promising for boosting the speed of computations and designs.
\section{Results}
\subsection{Metasurface Design}
Figure 1 demonstrates the proposed 8 annular models, each of them is composed of three layers of copper, a substrate (FR4 with permittivity of 4.2+0.025i, and thickness of 1.5mm), and a ground plane to block the incident electromagnetic waves.
Each annular model is 1.6 mm and is composed of $8 \times8$ lattices named as “1” and “0” to indicate the spaces with and without the copper. Each meta-atom consists of $4 \times4$ arbitrarily distributed annular models and the final unit-cell is 6.4 mm.
Therefore, each unit cell generated as an input of DNN comprises $32 \times32$ lattices with a length of 0.2 mm. The reason for using 8 annular patterns is to produce single or multiple resonances in the generated S-parameters in a broad frequency range from 4 to 45 GHz.
Since it is almost impossible to attain the relationship between the input ``0" and ``1" matrix and its equivalent S-parameter,  the machine learning method can be considered as an encouraging solution to shorten the computational operations for obtaining the ideal results.
 \renewcommand{\figurename}{Figure}
\begin{figure}
	\centering
	\includegraphics[height=7cm]{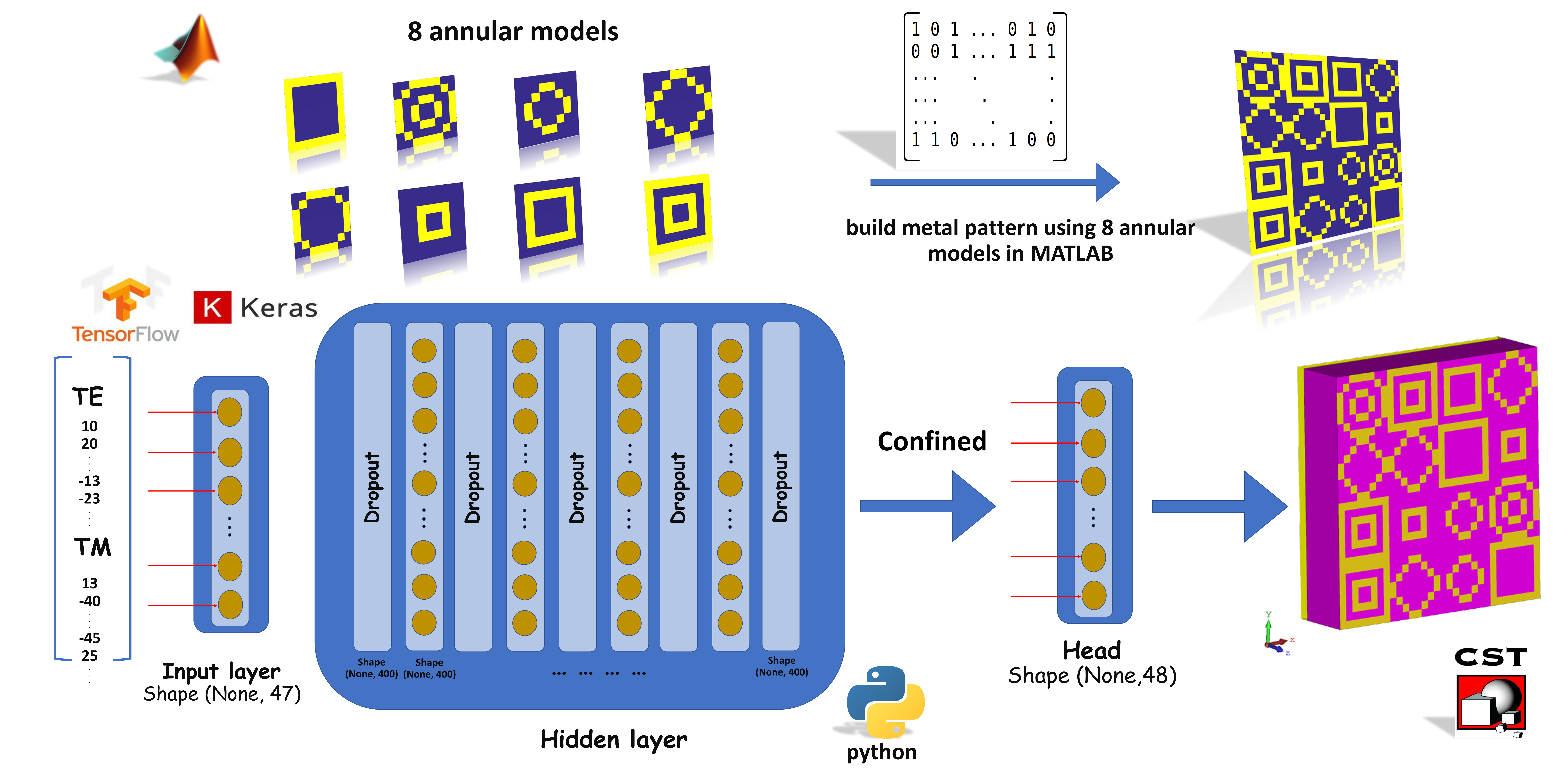}
	\caption{Diagram of the design procedure of confined- output configuration and 8 annular models.}
	\label{fgr:example2col}
\end{figure}
\subsection{Deep learning}
Nowadays, neural network algorithms have appeared to solve some fundamental challenges especially in optimization and artificial intelligence. Schematic representation of an artificial neuron is depicted in Figure 2, in which ${A_n}$ denotes the input neurons.
Since each $A$ is connected to a weight, the multiplication of $A_{i}$'s by $W_{i}$'s has emerged in the input of the summation. By considering  $\phi(x)$ as an activation function, eventually, the output of this process is determined by:
\begin{equation}
Y = \phi(\sum\limits_{i=1}^n W_{i}A_{i}+b_{i})
\end{equation}
In the above equation, $b$ indicates the bias value. Generally, artificial neural networks are made of distinct layers of input, output, and a hidden layer between them. By increasing the hidden layers, the complexity of the network increases, and the neural network turns into a deep neural network.
\begin{figure}
	\centering
	\includegraphics[height=6cm]{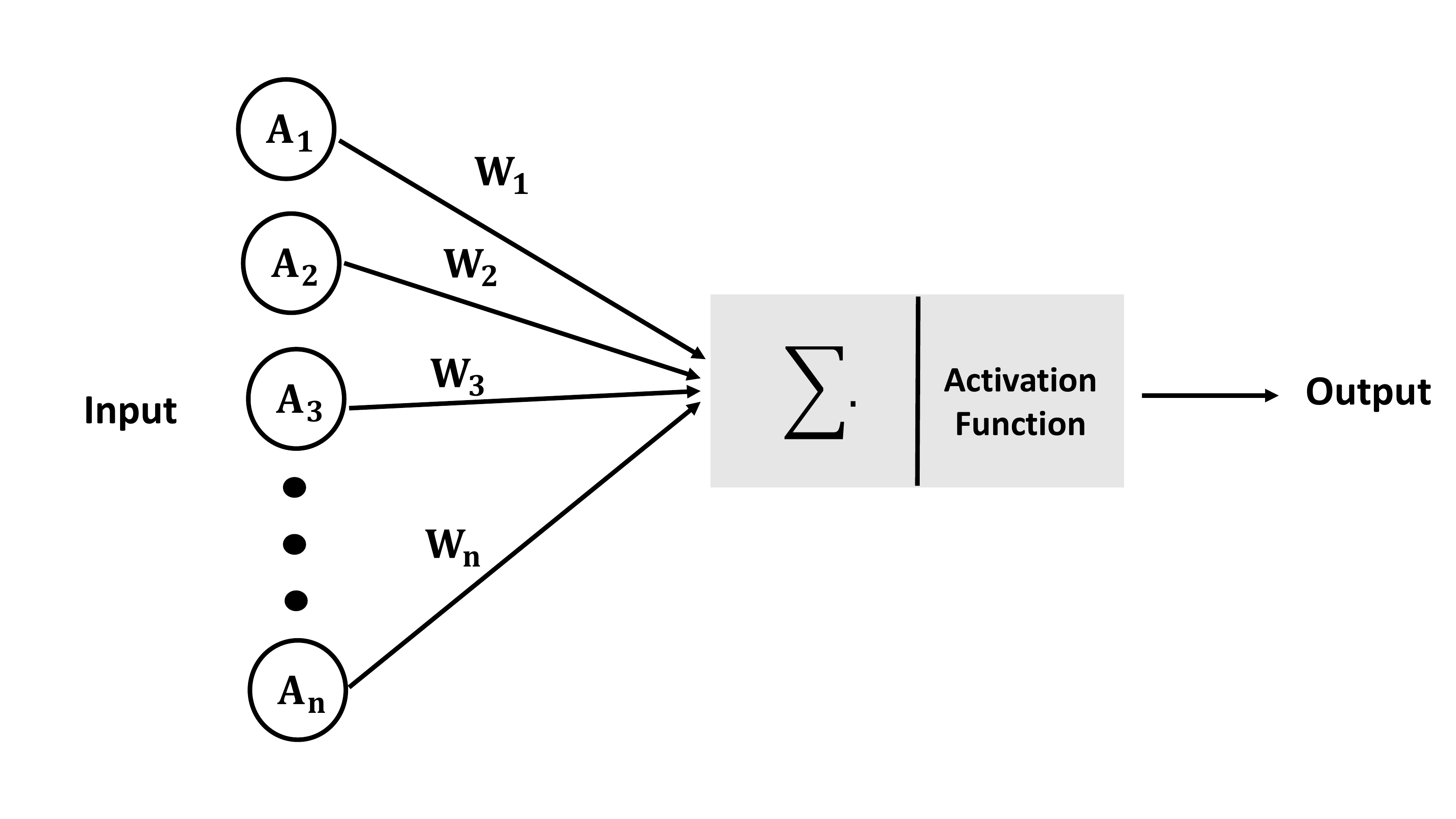}
	\caption{A sketch of an artificial neuron.}
	\label{fgr:example2col}
\end{figure}
\subsection{Confined output configuration}
In this paper, we have benefited from a DNN to find out the intrinsic connections between the output generated metasurface  and its S-parameters features. To establish our dataset, by the means of the RAND function,  a collection of 2000 matrices that form the unit cells are created. Then, we use CST Microwave Studio to determine the reflection characteristics of each unit cell under the illumination of both TE and TM polarized waves. Then, by linking CST MWS with MATLAB, the information regarding reflection characteristics (resonance frequency, resonance depth, and resonance bandwidth) are saved in a generated database. Note that for obtaining the S-parameters of the unit cells in periodic structures, periodic boundary conditions are adjusted along x- and y-directions, while open boundary conditions are applied along the propagation of the incoming waves.
 \renewcommand{\figurename}{Table}
  \renewcommand{\thefigure}{1}
\begin{figure*}
		\caption{Elaborate data of the confined output DNN configuration.}
	\centering
	\includegraphics[height=8cm]{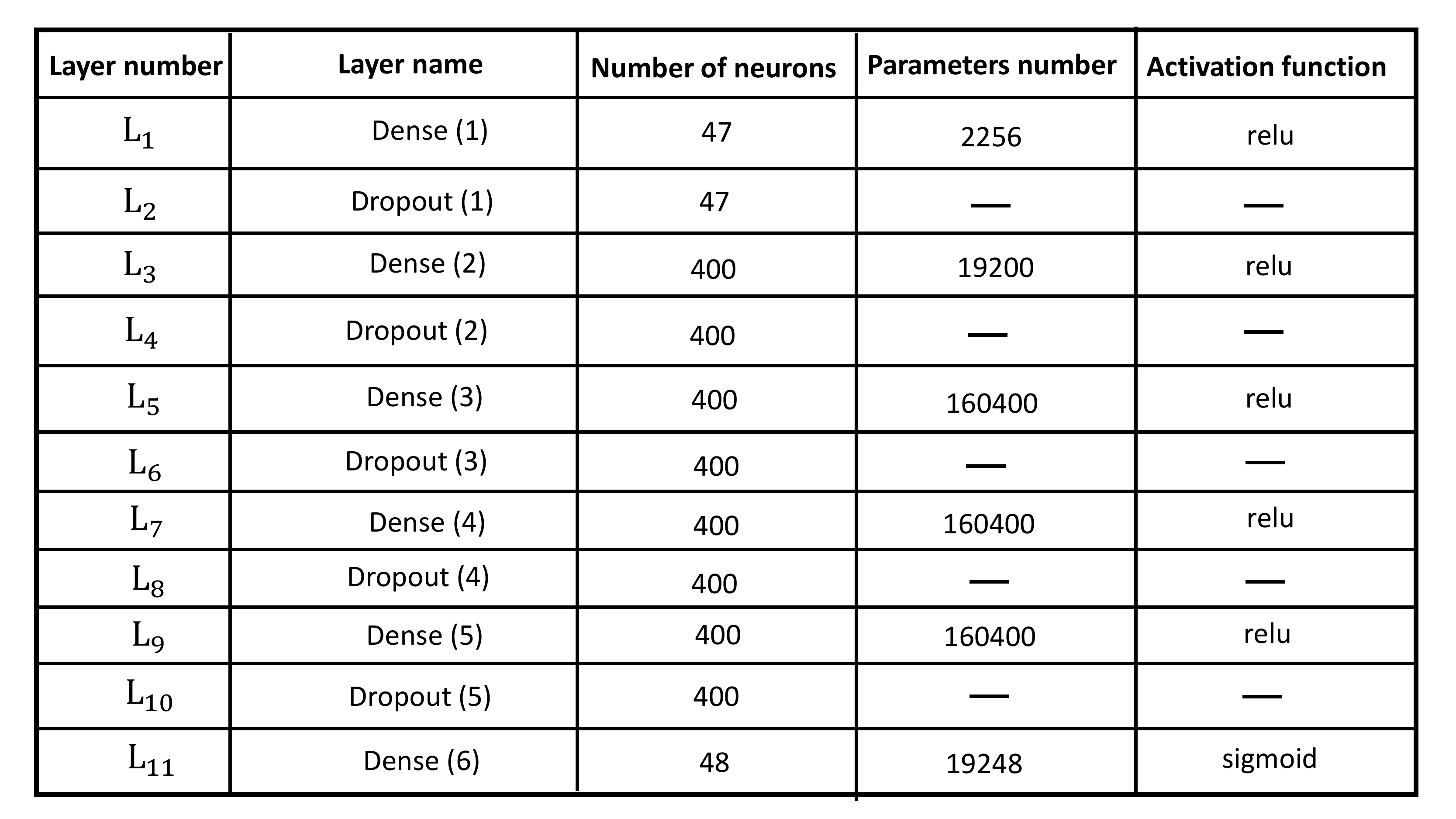}

	\label{fgr:example2col}
\end{figure*}
 \renewcommand{\figurename}{Figure}
\renewcommand{\thefigure}{3}
\begin{figure*}
	\centering
	\includegraphics[height=8cm]{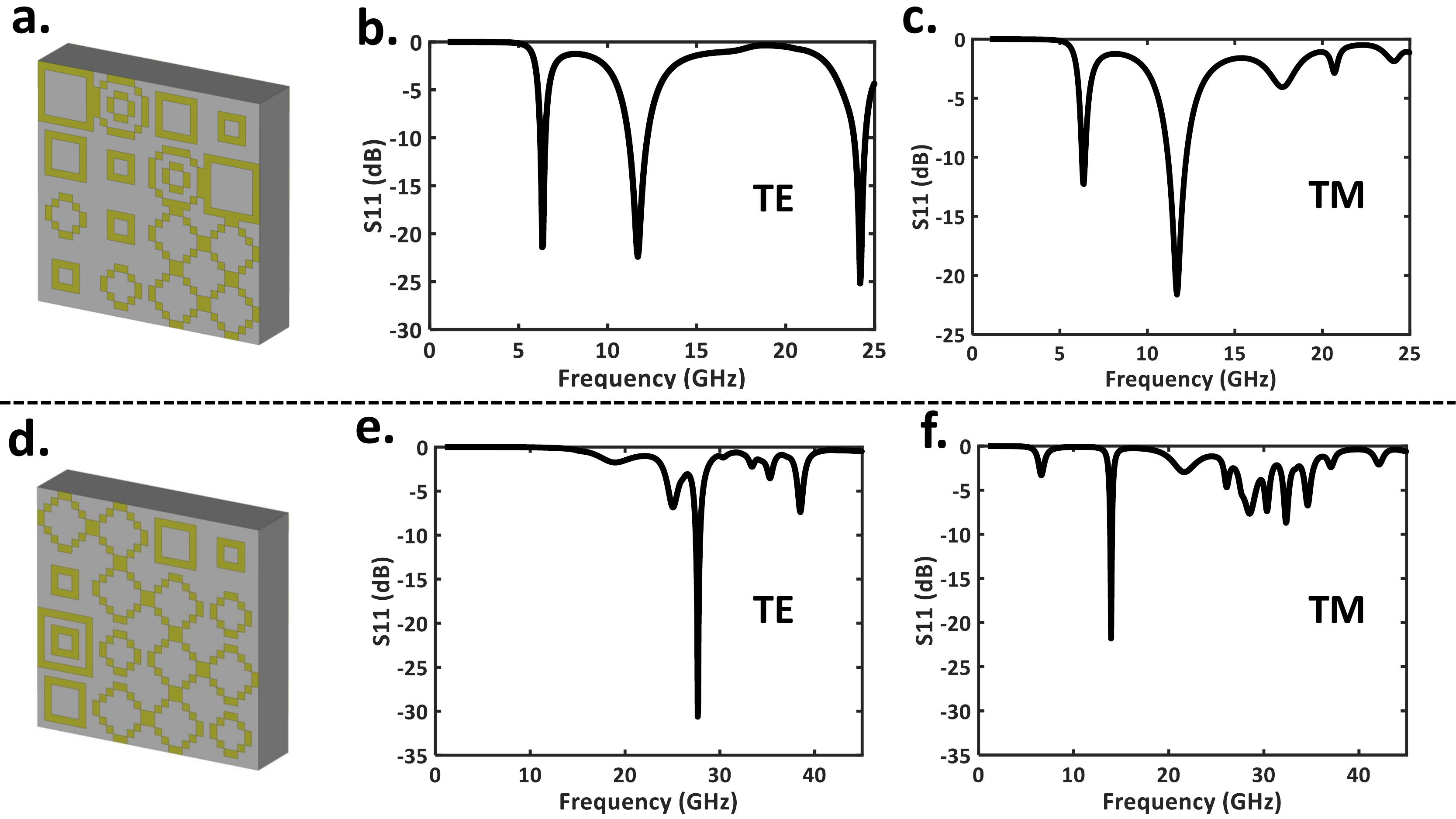}
	\caption{Two examples of reflection amplitude of confined output DNN configuration a,d) final metasurface structure. b,e) simulated S-parameters under illumination of TE and c,f) TM polarized waves.}
	\label{fgr:example2col}
\end{figure*}
 \renewcommand{\figurename}{Figure}
\renewcommand{\thefigure}{4}
\begin{figure*}
	\centering
	\includegraphics[height=8cm]{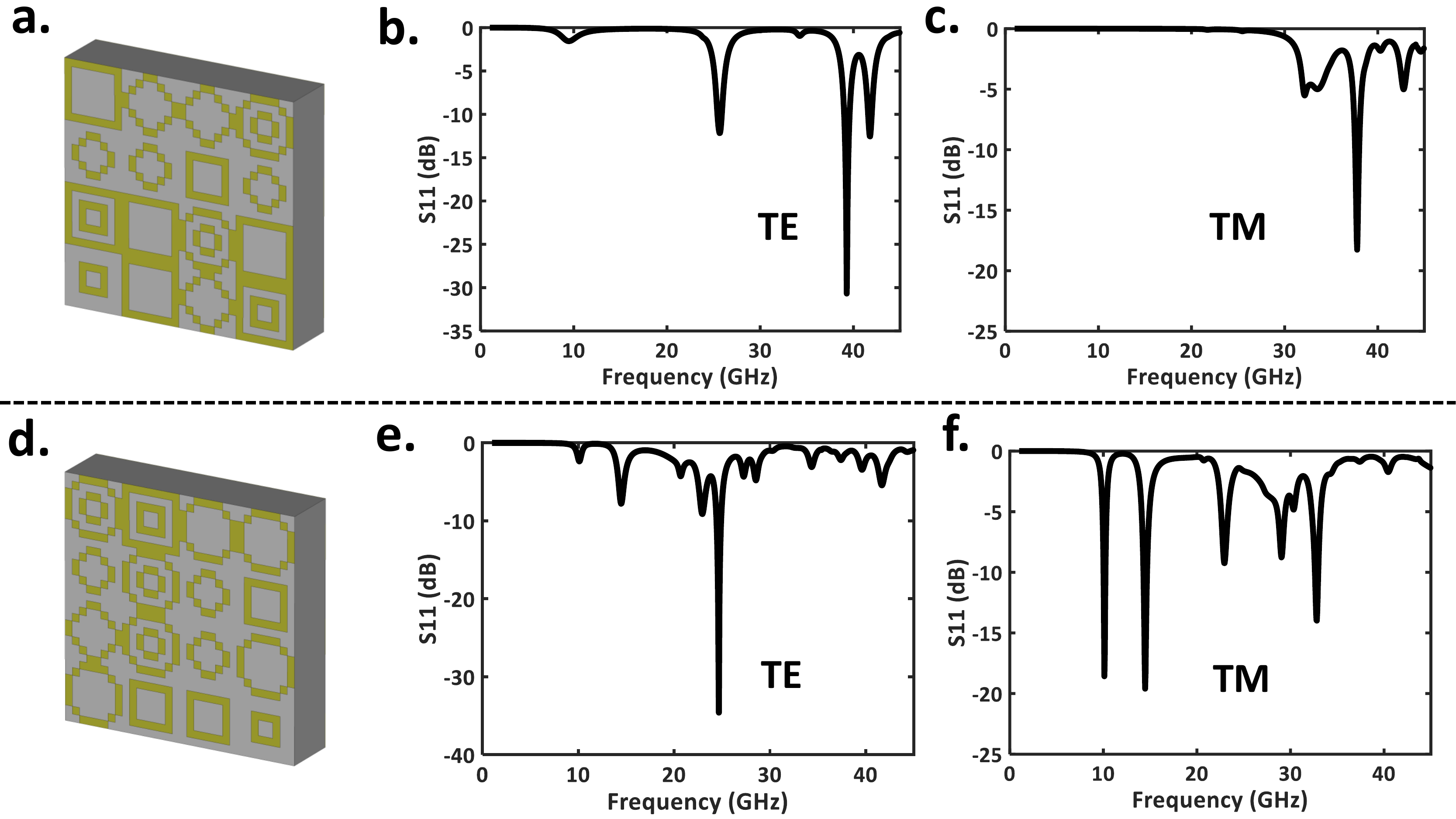}
	\caption{Two examples of reflection amplitude of confined output DNN configuration a,d) final metasurface structure. b,e) simulated S-parameters under illumination of TE and c,f) TM polarized waves.}
	\label{fgr:example2col}
\end{figure*}
\renewcommand{\thetable}{2}
\begin{table*}
	\centering
	\caption{Pre-determined TE and TM input targets (TE / TM) for desired S-parameters, that are demonstrated in Figure 3 and Figure 4.}
	\scalebox{0.9}{
		\begin{tabular}{c|c|c|c|c}
			\hline
			Examples & resonance frequency (GHz) & resonance depth (dB) & resonance bandwidth (GHz) \\ \hline
			Figure 3(a-c)                 & 6.5, 11.5, 24 / 6.5, 11.5    &-21.5, -22.5, -25 /-12, -21.5  & 0.3, 1, 0.6 / 0.2, 1                \\ \hline
			Figure 3(d-f)             & 27.5 / 14  & -30 / -22                    & 0.5 / 0.2                   \\ \hline
			Figure 4(a-c)             & 25.5, 40 / 38   & -12, -30 / -18            & 0.4, 0.6 / 0.5               \\ \hline
			Figure 4(d-f)            & 24.5 / 10, 14.5, 33  & -34.5 / -18.5, -20, -14 & 0.5 / 0.2, 0.4, 0.3                 \\ \hline
			
	\end{tabular}}
	
\end{table*}

Our dataset randomly generates 16 numbers from 1 to 8 to create $4 \times4$ matrix. Each number denotes one of the eight annular models; thus, the output of our established model is a matrix of $32 \times32$. In the training step, we have produced 2000 pairs of S-parameters (TE and TM) and their corresponding metasurface structures (70\% belongs to the training and the rest of the data belongs to a testing set). Each generated unit cell can produce up to eight resonances. Since we have extracted three features for each resonance (resonance frequency, resonance depth, and resonance bandwidth), a vector with a size of 24 has emerged in the input of our devised DNN. To enhance the network speed, minimize the volume of computations, and boost the effectiveness of our learning section, we have confined the output of the network; so that, the final metasurface is formed employing our designed 8 annular models. Note that to create the final vector, eight annular models are indicated by digital codes of “000” to “111”; thus, the output of the DNN originates a vector with the length of 48. This confined output configuration will reduce the volume of computations while maintaining the accuracy of the network. The details of the confined output configuration are outlined in Table I.

In our model, we have used dense and dropout layers successively as depicted in Figure 1. To escape from the misdirecting in the learning section and decreasing the chance of overfitting, we randomly neglected specified numbers of neurons in the dropout layer. On the other hand, we have defined the dense layer in such a way that each neuron has a connection to all the neurons in the former layers. Throughout the training section, we have used Adam optimizer to calculate the variances between the predetermined and output data repeatedly. By defining a loss function as a Mean Square Error (Eq. 2), we have observed the differences between the original and generated data; thus, when it reaches the specified criterion, the training step will stop. 

\begin{equation}
{\rm{MSE}} = \frac{1}{m}\sum\limits_{i = 1}^m {{{({Y_i} - {{\hat Y}_i})}^2}} 
\end{equation}
In the above equation, m, $Y_{i}$ and ${{{\hat Y}_i}}$ represent the number of data points, observed and predicted values, respectively. To increase the rate of accuracy, we adopted the sigmoid activation function in the layer number of 11 (see Table 1) since our wanted output in the DNN is 0 or 1. Eq. 3 and Eq. 4 show the expression of the activation of relu and sigmoid operators.
\begin{equation}
f(z) = \left\{ {\begin{array}{*{20}{l}}
	0&{for\,\,z < 0}\\
	z&{for\,\,z \ge 0}
	\end{array}} \right.
\end{equation}

\begin{equation}
\sigma (z) = \frac{1}{{1 + {e^{ - z}}}}
\end{equation}
When it comes to evaluating the model, different design targets of S-parameters are expected to determine whether the designed network can generate corresponding metasurface structures. To establish our model, the Tensorflow and Keras frameworks have been adopted, while our network is performed using Python  3.8.0. Finally, after the design steps are completed, it is only needed to insert the wanted S-parameter features, and the designed DNN can form the final unit cell in accordance with the learned information obtained through the training process.
\renewcommand{\thetable}{3}
\begin{table}
	\centering
	\caption{Detailed information of training and evaluation time, in addition to the model size for confined output DNN architecture.}
	\scalebox{1.1}{
		\begin{tabular}{c|c|c|c|c}
			\hline
			& Confined output network 
			\\ \hline
			Training time         & 19 minutes            \\ \hline
			Evaluation          & 0.038 sec          \\ \hline
			Model size          & 6 MB                              \\ \hline
			
	\end{tabular}}
	
\end{table}

To validate the efficiency of our presented DNN, four different examples are provided in the following. The output matrix of the metasurfaces is generated based on the input S-parameters data for both TE and TM polarized waves. The designated reflection information, which is detailed in Table II, is as follows for all the examples: [notches frequencies; notches depth; and notch bandwidth]. Numbers before and after the slash symbol ( / ) demonstrate the desired input data under the illumination of TE and TM-LP waves, respectively. 
Then, by entering the final generated matrices into full-wave simulation software, we obtained the simulated reflection amplitude under the illumination of dual orthogonal polarizations.
As an example, we intend to design a metasurface with an S-parameter containing three resonances under TE polarization while having two under -10 dB resonances under TM polarization waves in specified frequencies (see the first row of Table II). Observe in Figure 3(b,c) that the full-wave simulation successfully reaches the design targets.
\begin{figure}
	\centering
	\includegraphics[height=5cm]{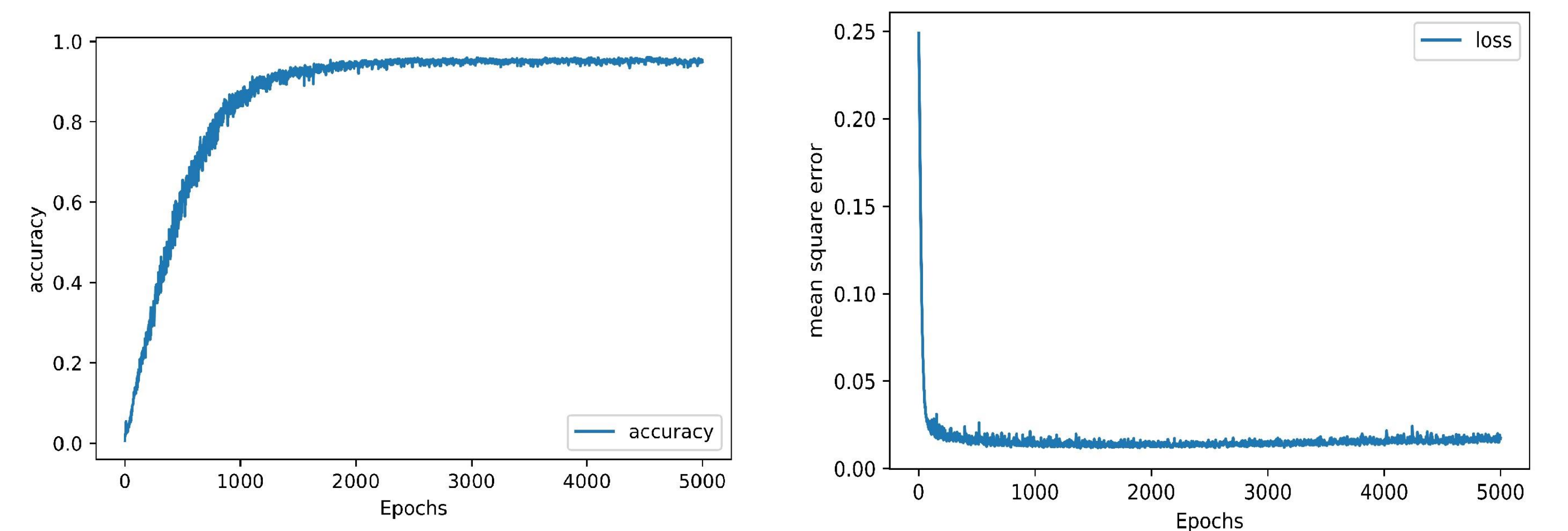}
	\caption{Diagrams of accuracy and mean square error according to 5000 Epochs for confined output DNN configuration.}
	\label{fgr:example2col}
\end{figure}

Similarly, for instance in the last example (see Figure 4(d-f)), a metasurface is designed in such a way that its S-parameters contain one and three resonances in pre-determined frequencies (see the last row of Table II). Full-wave simulation results are in accordance with our sought design goals. Moreover, the diagrams of the accuracy and loss function of our presented DNN confined output architecture are depicted in Figure 5. 
In addition, the details of training and evaluation time, and the model size for our proposed DNN are presented in Table III. These results are captured utilizing Google Colab with the model of Tesla T4 and 15 MB of RAM. As can be seen, the design time of our approach to generating a unit cell is 0.038 sec. Therefore, compared to the conventional method, which takes about 700 to 800 minutes, our presented approach is much faster and more efficient. Note that we cannot directly compare our method to other inverse designs of the metasurface since the employed GPU and RAM are different.

Accordingly, we have proven that our machine learning-based approach is a promising candidate for inverse design of the metasurfaces in terms of calculation repetitions, accuracy and speed. Establishing a DNN-confined output configuration based on the input data for both TE and TM polarized waves is the innovation to form a specified metasurface structure for dual orthogonal polarizations. 
\section{Conclusion}
In conclusion, adopting a deep neural network, we have presented an inverse metasurface design approach for dual orthogonal polarized waves. By merely specifying four design goals for both TE and TM cases (number of notches, notch frequencies, notch depths, and notch bandwidths), the designed DNN can create the final metasurface structure in the output. To broaden the working frequency, we have considered 8 annular models; so that the created unit cells in the output can produce different resonances over wide frequencies up to 45 GHz. The numerical simulations illustrate that our DNN can successfully generate the desired metasurface compared to our pre-determined designed targets, with an average accuracy higher than 91\%. 
We have shown that the speed of our presented approach is much higher than conventional metasurface design, and by proposing the confining output configuration, our approach equips an encouraging platform as an efficient technique with respect to computational repetitions, training and evaluation time, and average accuracy. We believe that our used deep neural network approach is a suitable candidate for inverse metasurface design for dual-polarized waves and complex wave-interaction phenomena.
\subsection{CONFLICT OF INTEREST}
The authors declare that there is no conflict of interest
\subsection{DATA AVAILABILITY}
The data that support the findings of this study are available
from the corresponding author upon reasonable request.

\end{document}